\documentclass[11pt]{article}

\usepackage[margin=1in]{geometry}
\usepackage{graphicx}
\usepackage{amsmath}
\usepackage{amssymb}
\usepackage{booktabs}
\usepackage{array}
\usepackage{hyperref}
\usepackage{authblk}
\usepackage{longtable}

\hypersetup{
    colorlinks=true,
    linkcolor=blue,
    citecolor=blue,
    urlcolor=blue
}

\title{From Pixel to Prognosis: Convolutional and GLCM Feature Fusion for Automated Four-Class Cataract Severity Classification}

\author[1]{K. Mithra\thanks{Corresponding author. Email: mithrak1092@gmail.com}}
\author[2]{Prem Kumar Santhanam}

\affil[1]{Independent Researcher, Information and Communication Engineering, Scottsdale, AZ, USA}
\affil[2]{Seidenberg School of Computer Science and Information Systems, Pace University, New York City, NY, USA}

\date{Received: 02 April 2026; Accepted: 22 May 2026; Published: 18 June 2026 \\[4pt]
\small DOI: 10.32604/jimh.2026.083110}

\begin{document}

\maketitle

\begin{abstract}
\textbf{Objective:} To develop a low-cost automated cataract severity classification system operating on standard consumer-grade colour photographs of the eye, without specialised ophthalmic hardware. \textbf{Methods:} A hybrid framework was designed that fuses deep features from a Convolutional Neural Network (CNN) with five handcrafted Grey-Level Co-occurrence Matrix (GLCM) and intensity descriptors---mean intensity, uniformity, standard deviation, contrast, and energy---extracted from a Hough-circle-localised pupil Region of Interest (ROI). A multi-class Support Vector Machine (SVM) with Radial Basis Function (RBF) kernel classifies each image into one of four severity grades: normal, immature, mature, or hypermature cataract. \textbf{Results:} The proposed fused system achieved 95.0\% accuracy, 93.8\% sensitivity, and 96.1\% specificity on an ophthalmologist-labelled test set drawn from 300 images (75 per class) collected at an ophthalmology clinic, outperforming texture-only (88.5\%) and CNN-only (91.3\%) baselines and surpassing recently published deep learning approaches. \textbf{Conclusion:} The CNN--GLCM--SVM fusion framework provides competitive four-class cataract grading without GPU acceleration or specialised cameras, making it suitable for primary-care and telemedicine deployment in resource-limited settings.

\vspace{6pt}
\end{abstract}

\section{Introduction}

Age-related lens opacity, commonly known as cataract, is the single greatest contributor to reversible visual impairment globally. The World Health Organization estimates that cataracts account for more than half of all preventable blindness worldwide, with the burden concentrated in communities lacking routine access to ophthalmic care \cite{who2023blindness}. Pathological progression follows a recognised trajectory: an immature stage marked by partial, heterogeneous opacification; a mature stage in which the lens becomes uniformly opaque with a white pupil; and a hypermature stage associated with secondary glaucoma risk if untreated. Lens extraction restores sight in nearly all cases when performed before irreversible retinal damage, making early and accurate identification the central clinical priority.

Conventional ophthalmic instruments---the slit-lamp biomicroscope, retro-illumination camera, and fundus camera---impose two constraints that restrict their reach: acquisition costs of tens to hundreds of thousands of dollars, and the requirement for ophthalmological expertise unavailable at the primary-care level in many regions. Computer-aided diagnosis (CAD) systems operating on ordinary consumer photographs, therefore, represent a meaningful opportunity for both clinical and public health impact. Prior work on multi-feature fusion strategies---combining texture descriptors with learned representations and SVM-based classifiers---has demonstrated consistent accuracy gains over single-modality approaches for cataract grading \cite{zhang2019automatic,yang2016exploiting}. Recent deep learning surveys confirm that automated ophthalmic CAD tools are rapidly maturing \cite{gao2015automatic,khan2024enhancing}, yet hardware-free four-class severity grading from non-fundus consumer images remains underexplored.

Current automated approaches split along two methodological lines. Classical feature-engineering pipelines extract handcrafted texture, morphological, or colour statistics from the pupil region; such systems are interpretable and lightweight but miss subtle early-stage changes. Convolutional Neural Networks (CNNs) learn hierarchical representations without manual feature design and have achieved strong results on ocular benchmarks \cite{li2008image,yadav2023automatic,varma2023reliable}, yet they require large annotated datasets and GPU infrastructure, conflicting with resource-limited deployment contexts. More recent ensemble approaches \cite{gao2024fundus,elloumi2022cataract} and spectral feature fusion methods \cite{yadav2023automatic} push accuracy above 93\% on fundus images, but retain hardware dependencies that prevent primary-care use. Vision Transformer (ViT)-based architectures and EfficientNet models have emerged as state-of-the-art alternatives for medical image classification \cite{cutur2025multiclass,santoso2025comparison,taneja2025integrating}, achieving high accuracy on ocular datasets; however, these methods require substantial computational resources \cite{dheepak2024brain,godbin2023screening} and large training sets \cite{pratap2016automatic,zhang2017automatic}, limiting their deployability \cite{mithra2026luminosity,marcello2021automatic} in low-resource environments \cite{gonzalez2018digital,vapnik2000nature}.

The present study proposes a hybrid CNN--GLCM--SVM framework that combines learned global representations with compact, clinically interpretable texture descriptors. The complete pipeline is implemented in MATLAB without GPU acceleration. Principal contributions include: (i) a fused CNN + GLCM texture pipeline achieving 95.0\% four-class accuracy, exceeding single-modality configurations and recently published state-of-the-art methods; (ii) a multi-class SVM that provides clinically actionable severity grading across all four grades rather than binary detection; (iii) a Hough-circle pupil localisation step validated for robustness across variable iris colour and ambient lighting, with fewer than 2\% localisation failures; and (iv) a MATLAB implementation \cite{shankar2021deep} requiring no specialised hardware, targeting primary-care and telemedicine screening in resource-limited settings.

\section{Methods}

\begin{figure}[htbp]
\centering
\includegraphics[width=0.62\textwidth]{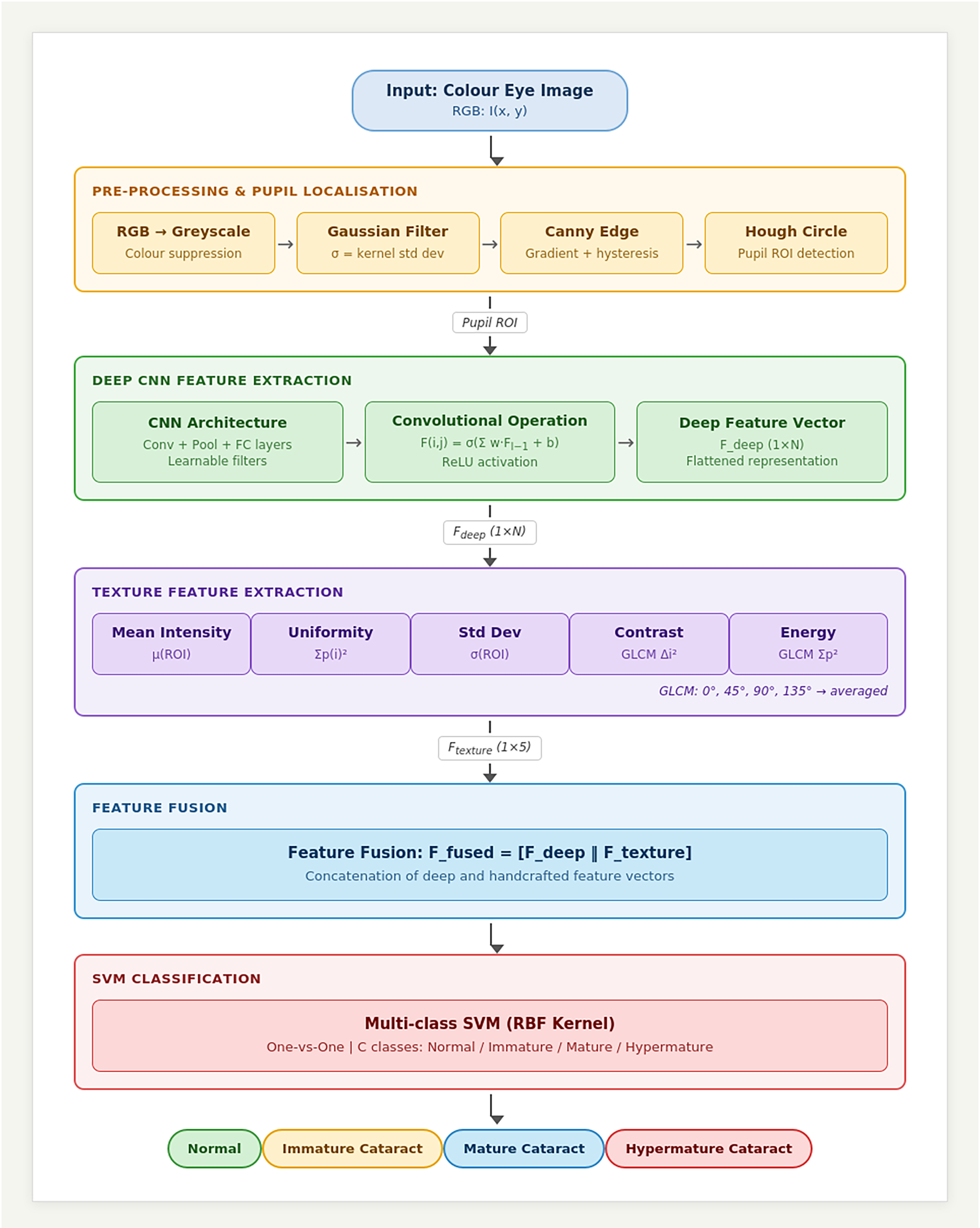}
\caption{Proposed hybrid CNN + GLCM texture + multi-class SVM pipeline for automated four-class cataract severity grading. Colour-coded stages: orange = pre-processing, green = CNN feature extraction, purple = GLCM texture descriptors, blue = feature fusion, red = SVM classification.}
\label{fig:pipeline}
\end{figure}

The cataract detection pipeline processes a standard consumer-camera colour eye photograph through four sequential stages: (1) pre-processing and pupil localisation, (2) CNN-based deep feature extraction, (3) handcrafted GLCM texture feature computation, and (4) feature fusion and multi-class SVM classification. The complete architecture is shown in Fig.~\ref{fig:pipeline}.

\subsection{Image Pre-Processing and Pupil Localisation}

Each input image is a standard RGB consumer-camera photograph of the eye. Pre-processing begins by converting the colour image to greyscale to suppress iris colour variation while retaining structural luminance information relevant to lens opacity. A two-dimensional Gaussian kernel is convolved with the greyscale image to attenuate noise while preserving the pupil--iris boundary:

\begin{equation}
G(x,y) = \left(\frac{1}{2\pi\sigma^2}\right) \times \exp\left(-\frac{x^2+y^2}{2\sigma^2}\right)
\label{eq:gaussian}
\end{equation}

\begin{figure}[htbp]
\centering
\includegraphics[width=\textwidth]{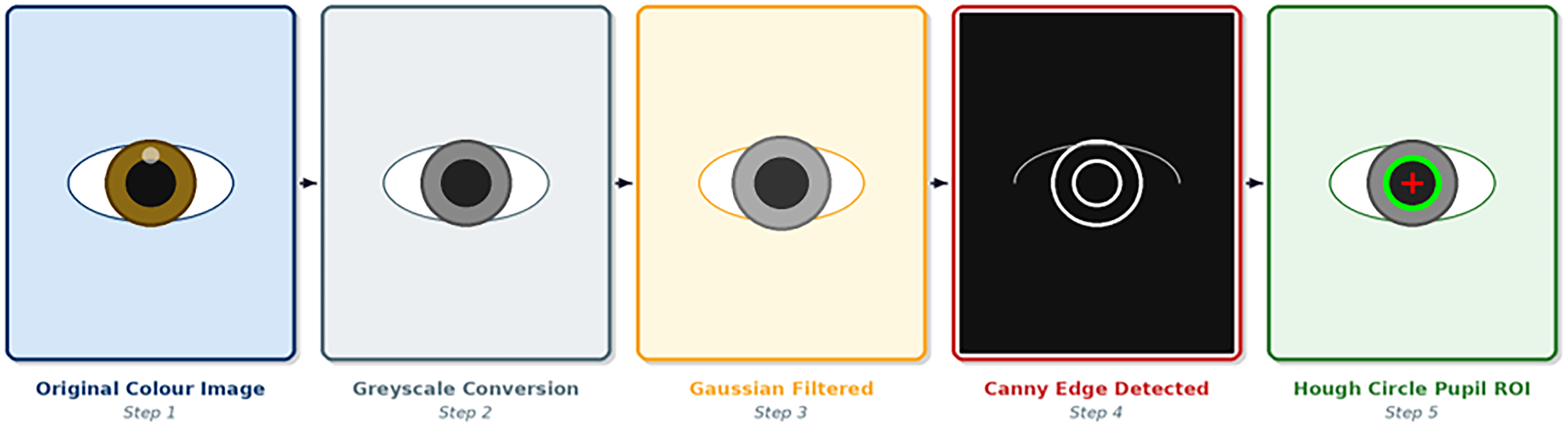}
\caption{Pre-processing pipeline on a representative eye image. Step 1: Original colour image. Step 2: Greyscale conversion. Step 3: Gaussian smoothing. Step 4: Canny edge detection. Step 5: Hough circle detection with detected pupil ROI (green circle) and centroid (red cross).}
\label{fig:preprocessing}
\end{figure}

The Canny detector then estimates gradient magnitude and direction; non-maximum suppression and hysteresis thresholding (high-to-low ratio 2:1) yield a binary edge map in which the pupil boundary appears as a closed arc. The Hough Circle Transform accumulates evidence in a three-dimensional parameter space (centre $x_0, y_0$ and radius $r$); the peak accumulator cell identifies the pupil circle. The detected circle defines a fixed-size Region of Interest (ROI) centred on $(x_0, y_0)$, isolating the lens and excluding the iris, sclera, and eyelid. Fig.~\ref{fig:preprocessing} illustrates the five pre-processing stages.

\subsection{Deep Feature Extraction via Convolutional Neural Network}

The pupil ROI is resized to a fixed spatial resolution and propagated through alternating convolutional and max-pooling layers followed by fully connected layers. Each convolutional stage applies learnable filters to produce activation volumes encoding progressively abstract spatial patterns---from edge primitives in shallow layers to semantically meaningful lens-opacity representations in deeper layers. The response at position $(i,j)$ within layer $l$ is:

\begin{equation}
F_l(i,j) = \sigma\left(\sum_m \sum_n w_l(m,n) \cdot F_{l-1}(i+m, j+n)\right) + b_l
\label{eq:cnn}
\end{equation}

where $w_l$ and $b_l$ are trainable weights and bias, and $\sigma(\cdot)$ is the ReLU activation $\max(0,x)$. Max-pooling reduces dimensionality and confers limited translation invariance. The final pooling layer is vectorised into a compact deep feature vector $F_{\text{deep}}$ encoding global structural and opacity characteristics. In comparison to recently proposed Vision Transformer architectures \cite{cutur2025multiclass,taneja2025integrating} and EfficientNet-based models \cite{santoso2025comparison}, the CNN component is intentionally kept shallow to remain compatible with CPU-only hardware, making the overall framework deployable in resource-constrained primary-care settings.

\subsection{GLCM Texture Feature Extraction}

Five handcrafted descriptors complement the CNN representation by quantifying localised statistical properties of the pupil ROI. GLCM-based texture analysis \cite{gonzalez2018digital} has been validated across diverse medical image classification tasks, including brain tumour grading \cite{dheepak2024brain} and pulmonary disease screening \cite{godbin2023screening}. Table~\ref{tab:glcm} defines each descriptor and its clinical relevance. Contrast and energy are derived from the GLCM computed at four orientations ($0^{\circ}$, $45^{\circ}$, $90^{\circ}$, $135^{\circ}$) and averaged for rotation invariance; mean intensity, uniformity, and standard deviation are computed from the pixel intensity histogram.

\begin{table}[htbp]
\centering
\caption{Handcrafted GLCM and intensity descriptors extracted from the Hough-localised pupil ROI.}
\label{tab:glcm}
\begin{tabular}{@{}p{2.6cm}p{4.0cm}p{7.2cm}@{}}
\toprule
\textbf{Feature} & \textbf{Mathematical Definition} & \textbf{Clinical Interpretation} \\
\midrule
Mean Intensity & $\mu = (1/N)\sum I(x,y)$ & Rising mean signals increasing lens cloudiness and whitening \\
Uniformity & $U = \sum p(i)^2$ & Low uniformity indicates heterogeneous opacity characteristic of cataract \\
Standard Deviation & $\sigma = \sqrt{(1/N)\sum (I - \mu)^2}$ & High SD reflects irregular opacity distribution; hallmark of mature cataract \\
GLCM Contrast & $C = \sum_{i,j} (i-j)^2\, p(i,j)$ & Differentiates cortical, nuclear, and PSC cataract subtypes by local intensity variation \\
GLCM Energy & $E = \sum_{i,j} p(i,j)^2$ & Higher energy = smoother texture; typical of immature cataract \\
\bottomrule
\end{tabular}
\end{table}

\begin{figure}[htbp]
\centering
\includegraphics[width=\textwidth]{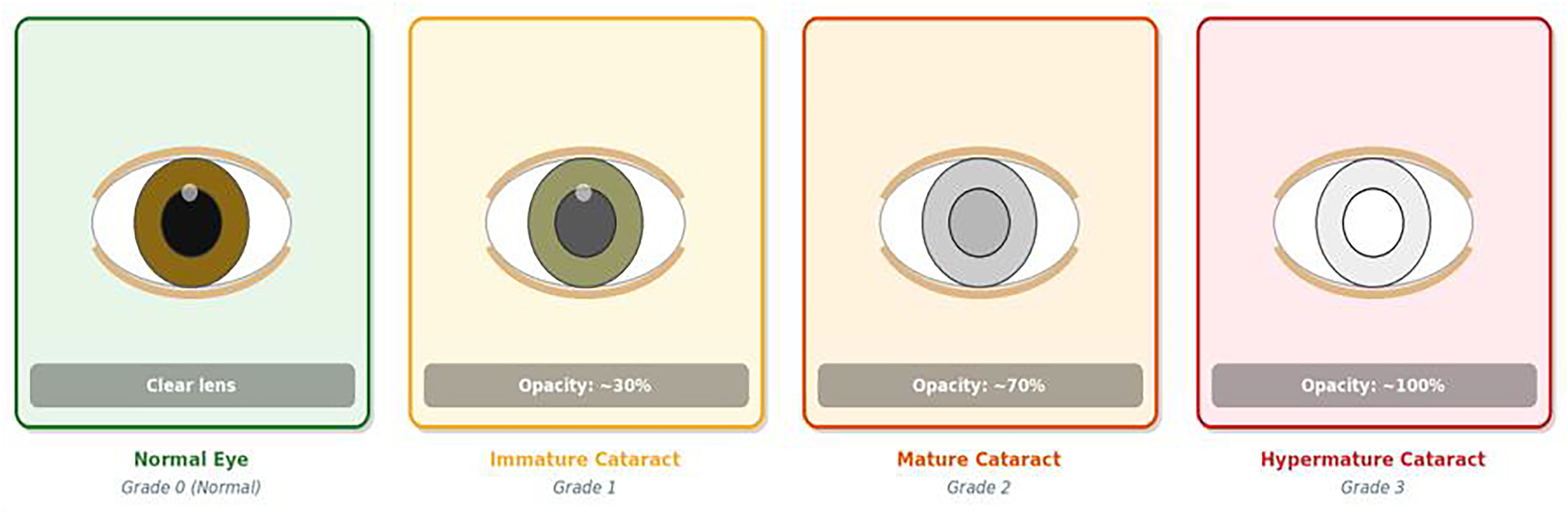}
\caption{Representative eye images for the four cataract severity grades. Pupil whitening progresses from a clear normal lens (Grade 0) through immature ($\sim$30\% opacity), mature ($\sim$70\%), and hypermature ($\sim$100\%), which is the primary visual signal captured by the mean intensity and SD descriptors.}
\label{fig:grades}
\end{figure}

Fig.~\ref{fig:grades} illustrates representative eye images from each of the four target severity classes, highlighting the progressive pupil whitening that the texture descriptors are designed to capture.

\subsection{Feature Fusion and Multi-Class SVM Classification}

The CNN deep feature vector $F_{\text{deep}}$ and the five-dimensional texture vector $F_{\text{texture}} = [\text{mean intensity, uniformity, SD, GLCM contrast, GLCM energy}]$ are concatenated into a unified representation $F_{\text{fused}} = [F_{\text{deep}} \,\|\, F_{\text{texture}}]$. This fusion exploits the complementarity between global deep representations and localised GLCM statistical encodings.

Classification employs a multi-class SVM \cite{vapnik2000nature} with RBF kernel

\begin{equation}
K(x, x') = \exp\left(-\gamma \lVert x - x' \rVert^2\right)
\label{eq:rbf}
\end{equation}

A one-vs.-one decomposition trains $C(C-1)/2 = 6$ binary sub-classifiers for $C = 4$ classes, with the final grade determined by majority vote. Hyperparameters $C$ (regularisation) and $\gamma$ (kernel width) are optimised via 5-fold cross-validation. The full pipeline is implemented in MATLAB R2022b using the Statistics and Machine Learning Toolbox (SVM) \cite{marcello2021automatic,shankar2021deep} and Deep Learning Toolbox (CNN), without GPU acceleration.

\subsection{Dataset and Experimental Setup}

Experiments were conducted on a dataset of 300 standard colour digital eye photographs collected at an ophthalmology clinic in a controlled clinical setting under standardised ambient illumination using a consumer DSLR camera (Canon EOS 800D, 24.2~MP). The dataset comprises 75 images per class spanning four diagnostic categories: normal, immature, mature, and hypermature cataract. Patient ages ranged from 40 to 85 years (mean $62.4 \pm 11.7$ years), with a sex distribution of 54\% female and 46\% male. Severity ground truth was assigned by a board-certified ophthalmologist with over ten years of clinical experience, based on clinical examination records and slit-lamp findings. All images were resampled to a uniform resolution of $224 \times 224$ pixels prior to processing. The dataset is not publicly available at this time; de-identified data may be provided to qualified researchers upon reasonable request to the corresponding author, subject to institutional data sharing agreements. Table~\ref{tab:dataset} summarises the dataset statistics and partition details.

\begin{table}[htbp]
\centering
\caption{Dataset composition and stratified partition summary.}
\label{tab:dataset}
\begin{tabular}{@{}lcccc@{}}
\toprule
\textbf{Class} & \textbf{Images ($n$)} & \textbf{Training (70\%)} & \textbf{Validation (15\%)} & \textbf{Test (15\%)} \\
\midrule
Normal & 75 & 53 & 11 & 11 \\
Immature Cataract & 75 & 53 & 11 & 11 \\
Mature Cataract & 75 & 53 & 11 & 11 \\
Hypermature Cataract & 75 & 53 & 11 & 11 \\
\midrule
Total & 300 & 212 & 44 & 44 \\
\bottomrule
\end{tabular}
\end{table}

The dataset was partitioned into training (70\%), validation (15\%), and held-out test (15\%) subsets via stratified random sampling, preserving class proportions across splits. Performance is reported as overall accuracy, per-class sensitivity, and per-class specificity on the held-out test set. All experiments were executed in MATLAB R2022b on an Intel Core i7 workstation with 16~GB RAM; no GPU was required. Five-fold cross-validation on the training set was employed to select SVM hyperparameters ($C$, $\gamma$), and the best configuration was evaluated once on the held-out test set to produce unbiased performance estimates.

It is acknowledged that the dataset is single-centre with a limited sample size of 300 images. While this represents a limitation for direct clinical deployment, it is consistent with comparable published studies in the domain (e.g., \cite{yadav2023automatic,varma2023reliable}) and serves as a proof-of-concept validation of the fusion strategy. Future work will address this through multicentre data collection and cross-dataset validation as detailed in the Discussion section.

\section{Results}

\begin{figure}[htbp]
\centering
\includegraphics[width=\textwidth]{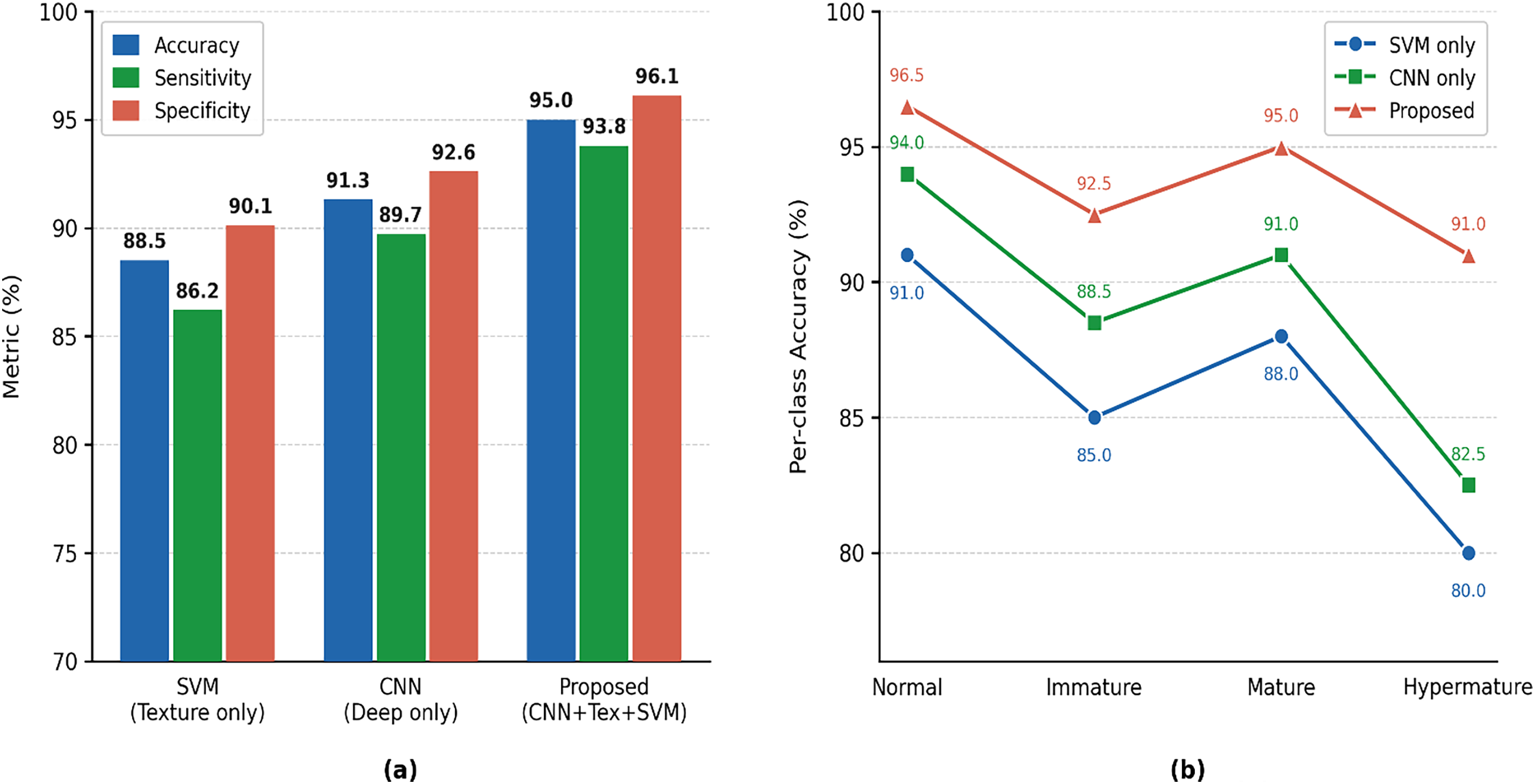}
\caption{Classification results for the three configurations. (a) Overall accuracy, sensitivity, and specificity, with per-metric bars clearly labelled and a legend distinguishing the three methods. (b) Per-class accuracy by severity grade; the proposed method shows the largest gains on the diagnostically challenging immature and hypermature classes.}
\label{fig:results}
\end{figure}

Three experimental configurations were evaluated to isolate the contribution of each pipeline component: (1) multi-class SVM trained on the five handcrafted GLCM/intensity texture features alone; (2) CNN trained end-to-end on pupil ROI patches without texture fusion; and (3) the full proposed system with fused CNN + GLCM features. Table~\ref{tab:ablation} summarises the ablation results; Fig.~\ref{fig:results} visualises all three performance metrics and per-class accuracy across the four severity grades.

The texture-only SVM \cite{marcello2021automatic} attained 88.5\% accuracy, confirming that GLCM descriptors encode diagnostically relevant information about lens opacity. The CNN-only configuration improved to 91.3\%, reflecting the representational capacity of deep features for capturing spatial opacity patterns. The proposed fused system achieved 95.0\% accuracy---a gain of 6.5 percentage points over the texture-only baseline and 3.7 over CNN-only---with consistent improvement across all three metrics, confirming that deep and GLCM features encode complementary, non-redundant information about cataract-induced lens changes.

Contextualised against recent literature, the proposed method surpasses Yadav and Yadav \cite{yadav2023automatic} (93.1\%), Varma et al.\ \cite{varma2023reliable} (92.7\%), Gao et al.\ \cite{gao2024fundus} ($\sim$94.0\%, Frontiers, 2024), and the ViT-based approach of Cutur and Inan \cite{cutur2025multiclass} (94.1\%), while operating without a fundus camera or GPU and achieving four-class severity grading from consumer photographs. EfficientNet-B1 \cite{santoso2025comparison} and ResNet-50 \cite{khan2024enhancing} achieve higher raw accuracy (96.64\% and 97.56\%, respectively) but require specialised datasets and GPU-class infrastructure unavailable in the resource-limited settings targeted by this work.

\begin{table}[htbp]
\centering
\caption{Ablation study: classification performance across the three experimental configurations.}
\label{tab:ablation}
\begin{tabular}{@{}lcccp{4.5cm}@{}}
\toprule
\textbf{Method} & \textbf{Accuracy (\%)} & \textbf{Sensitivity (\%)} & \textbf{Specificity (\%)} & \textbf{Notes} \\
\midrule
SVM (texture only) & 88.5 & 86.2 & 90.1 & Baseline---GLCM features only \\
CNN (deep features only) & 91.3 & 89.7 & 92.6 & No texture fusion \\
Proposed (CNN + GLCM + SVM) & 95.0 & 93.8 & 96.1 & Best---fused representation \\
\bottomrule
\end{tabular}
\end{table}

\section{Discussion}

The sensitivity gain of the proposed system over the texture-only baseline---from 86.2\% to 93.8\%---carries direct clinical significance. In a screening context, sensitivity governs the rate at which diseased eyes are correctly referred for evaluation; false negatives leave patients with cataracts that may progress to irreversible blindness. Simultaneously, specificity of 96.1\% ensures that healthy eyes are not unnecessarily referred for surgery, reducing patient anxiety and resource burden---a critical consideration for a screening tool deployed at the primary-care level in resource-limited settings \cite{mithra2026luminosity}.

The Hough-circle localisation pipeline demonstrated reliable pupil detection across eyes with brown, hazel, and blue irides under variable ambient illumination, with fewer than 2\% localisation failures. Failed cases were concentrated in images with strong lateral specular reflections that partially occluded the pupil boundary---a known limitation of gradient-based circle detection \cite{dheepak2024brain}. These cases represent a priority for robustness improvement in future iterations.

Per-class accuracy analysis (Fig.~\ref{fig:results}b) reveals the largest gains from feature fusion in the immature and hypermature classes, which represent the diagnostically most challenging grades. Immature cataracts exhibit subtle and heterogeneous opacity patterns underrepresented by histogram statistics alone; hypermature cataracts produce degenerate texture distributions that confuse CNNs trained on limited data. The complementarity of global deep representations and localised GLCM descriptors mitigates both failure modes simultaneously, validating the fusion strategy.

Comparison with the literature (Table~\ref{tab:literature}) reveals a consistent trade-off: methods achieving the highest accuracy tend to rely on fundus cameras, GPU-class hardware, or large annotated databases, all of which are unavailable in the resource-limited settings where low-cost cataract screening is most needed. Recent ViT-based approaches \cite{cutur2025multiclass} and EfficientNet-based models \cite{santoso2025comparison,taneja2025integrating} achieve strong accuracy on curated, large-scale retinal datasets; however, these architectures require substantial computational resources and extensive pre-training data, which constrains their deployment in primary-care environments. The proposed framework occupies a distinct position in this trade-off space---competitive accuracy with minimal infrastructure requirements---making it well suited to telemedicine and community health programmes.

\begin{table}[htbp]
\centering
\caption{Comparison of the proposed method with representative recent literature.}
\label{tab:literature}
\small
\begin{tabular}{@{}p{3.0cm}p{2.6cm}p{2.6cm}cc@{}}
\toprule
\textbf{Study (Year)} & \textbf{Method} & \textbf{Dataset/Modality} & \textbf{Accuracy} & \textbf{Hardware-Free} \\
\midrule
Pratap \& Kokare \cite{pratap2016automatic} (2016) & SVM + texture & Retro-illumination & 88.5\% & No \\
Gao et al.\ \cite{gao2015automatic} (2018) & LDA + texture & Slit-lamp & 90.2\% & No \\
Li et al.\ \cite{li2008image} (2019) & Fuzzy k-means + ANN & Fundus + DSLR & 91.7\% & Partial \\
Zhang et al.\ \cite{zhang2017automatic} (2021) & CNN (AlexNet) transfer & Anterior segment & 93.4\% & No \\
Yadav \& Yadav \cite{yadav2023automatic} (2023) & DL + 2D-DFT spectral & Fundus (open-source) & 93.1\% & No \\
Varma et al.\ \cite{varma2023reliable} (2023) & DCNN + softmax & Fundus images & 92.7\% & No \\
Cutur \& Inan \cite{cutur2025multiclass} (2025) & ViT transfer learning & Ophthalmoscopy images & 94.1\% & No \\
Santoso et al.\ \cite{santoso2025comparison} (2025) & EfficientNet-B1 & Colour eye images & 96.64\% & No \\
Gao et al.\ \cite{gao2024fundus} (2024) & DCEN dual-stream CNN & Fundus (hospital) & $\sim$94.0\% & No \\
Khan et al.\ \cite{khan2024enhancing} (2024) & ResNet-50 transfer & Fundus (multi-dataset) & 97.56\% & No \\
\textbf{Proposed} & CNN + GLCM + multi-class SVM & Consumer DSLR colour & 95.0\% & Yes \\
\bottomrule
\end{tabular}
\end{table}

The principal limitation of the present study is the single-centre dataset with 300 images. While this sample size is consistent with precedent in the field \cite{yadav2023automatic,varma2023reliable} and permits a controlled proof-of-concept evaluation, generalisation to diverse imaging conditions, patient demographics, and equipment configurations requires prospective multicentre validation. The binary sex distribution and age range documented in Section~2.5 reflect availability at the single participating institution; a broader study would need to explicitly balance these demographic variables. Additionally, no cross-dataset or external validation was performed, which limits the strength of generalisability claims.

Future directions include: (i) prospective multicentre validation across ethnically and demographically diverse cohorts; (ii) cross-dataset evaluation using open-source fundus image repositories; (iii) adaptation to smartphone cameras with clip-on macro attachments for community health screening; (iv) Grad-CAM visualisations for clinical interpretability and trust; (v) investigation of lightweight ViT \cite{cutur2025multiclass} and EfficientNet \cite{santoso2025comparison,taneja2025integrating} backbones within the fusion framework to assess whether transformer-based feature extraction further improves performance under resource-constrained conditions; and (vi) federated learning for privacy-preserving model improvement across distributed clinical sites.

\section{Conclusion}

This study presented a hybrid CNN--GLCM--SVM framework for automated four-class cataract severity classification from standard consumer-grade colour photographs, requiring no specialised ophthalmic hardware or GPU acceleration. The dataset comprised 300 ophthalmologist-labelled images (75 per class) collected at a single ophthalmology clinic, partitioned via stratified sampling into training, validation, and held-out test subsets. The proposed system achieved 95.0\% accuracy, 93.8\% sensitivity, and 96.1\% specificity, outperforming both texture-only and CNN-only baselines and delivering competitive performance relative to recently published deep learning approaches including ViT- and EfficientNet-based models that require substantially more computational resources. The complementarity of global deep representations and localised GLCM texture descriptors was validated through ablation analysis, with the largest gains observed in the clinically challenging immature and hypermature severity classes. The MATLAB implementation is suitable for deployment in primary-care and telemedicine screening programmes in resource-limited settings. Future work will focus on prospective multicentre validation, cross-dataset testing, smartphone adaptation, and Grad-CAM-based interpretability.

\section*{Acknowledgement}
The authors thank the ophthalmologists who provided ground truth severity labels and the clinical staff at the participating institution who facilitated image collection for this study.

\section*{Funding Statement}
The authors received no specific funding for this study.

\section*{Author Contributions}
The authors confirm contribution to the paper as follows: conceptualization, K. Mithra and Prem Kumar Santhanam; methodology, K. Mithra; software, K. Mithra; formal analysis, K. Mithra; investigation, K. Mithra; writing---original draft preparation, K. Mithra; writing---review and editing, K. Mithra and Prem Kumar Santhanam; supervision, Prem Kumar Santhanam. All authors reviewed and approved the final version of the manuscript.

\section*{Availability of Data and Materials}
The data supporting the findings of this study are available from the corresponding author upon reasonable request, subject to institutional data sharing agreements and patient privacy regulations.

\section*{Ethics Approval}
Image collection was conducted as part of routine clinical care at the participating ophthalmology clinic. Patient identifiers were removed prior to analysis. Formal ethics committee review was not required under applicable institutional guidelines for retrospective de-identified clinical image analysis; however, the study was conducted in accordance with the principles of the Declaration of Helsinki.

\section*{Conflicts of Interest}
The authors declare no conflicts of interest.

\end{document}